\documentclass[sigconf,authorversion,nonacm]{acmart}

\usepackage[normalem]{ulem}
\useunder{\uline}{\ul}{}
\usepackage{subfig}

\AtBeginDocument{%
  \providecommand\BibTeX{{%
    \normalfont B\kern-0.5em{\scshape i\kern-0.25em b}\kern-0.8em\TeX}}}

\begin{document}

\title{Rating Prediction in Conversational Task Assistants 
with Behavioral and Conversational-Flow Features
}


\author{Rafael Ferreira}
\affiliation{%
  \institution{NOVA University of Lisbon \\ NOVA LINCS}
  \city{Lisbon}
  \country{Portugal}}
\email{rah.ferreira@campus.fct.unl.pt}

\author{David Semedo}
\affiliation{%
  \institution{NOVA University of Lisbon \\ NOVA LINCS}
  \city{Lisbon}
  \country{Portugal}}
\email{df.semedo@fct.unl.pt}

\author{João Magalhães}
\affiliation{%
  \institution{NOVA University of Lisbon \\ NOVA LINCS}
  \city{Lisbon}
  \country{Portugal}}
\email{jm.magalhaes@fct.unl.pt}

\renewcommand{\shortauthors}{Rafael Ferreira, David Semedo and João Magalhães}

\begin{abstract}
Predicting the success of Conversational Task Assistants (CTA) can be critical to understand user behavior and act accordingly.
In this paper, we propose TB-Rater, a Transformer model which combines conversational-flow features with user behavior features for predicting user ratings in a CTA scenario. 
In particular, we use real human-agent conversations and ratings collected in the Alexa TaskBot challenge, a novel multimodal and multi-turn conversational context.
Our results show the advantages of modeling both the conversational-flow and behavioral aspects of the conversation in a single model for offline rating prediction.
Additionally, an analysis of the CTA-specific behavioral features brings insights into this setting and can be used to bootstrap future systems.

\end{abstract}


%

\maketitle

\section{Introduction}
\label{sec_intro}
\begin{figure}[t]
    \centering
    \includegraphics[width=0.90\linewidth]{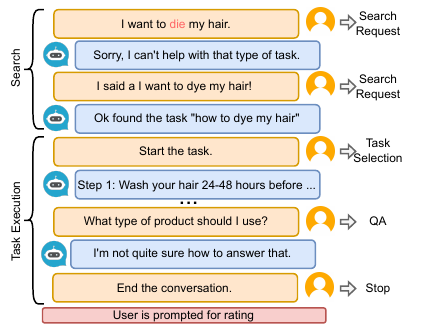}
    \caption{User and CTA example of a low-rated interaction. System and user utterances were emulated from real dialogs.
    }
    \label{fig_bad_conversation}
\end{figure}

Recently, Conversational Task Assistants (CTA) that are able to guide users through manual tasks are gathering more attention~\cite{taskbot_overview, task2dial, wizard_of_tasks} due to their applicability in everyday routines.
These differ and expand from other paradigms, such as conversational search~\cite{conversational_search} and task-oriented conversational agents~\cite{advances_in_tod}. 
In these paradigms, the user provides information to the assistant, and the system performs a task (e.g., searching or buying a ticket). In a CTA setting, it is the user that completes a task with the help of an assistant~\cite{taskbot_overview}. 
Creating these assistants requires various sub-systems working hand-in-hand to effectively help the user complete a variety of tasks such as ``baking a cake'' or ``fixing a leaky faucet''~\cite{taskbot_overview}.
Figure~\ref{fig_bad_conversation} illustrates a partial CTA dialog. First, users are prompted to search for the task they want to do, which is done using an IR step.
Second, the user selects one of the provided tasks and enters the task-execution phase. 
Third, the task instructions are presented to the user.
The user is then able to follow the task or create conversational sub-flows by asking task-specific or general questions, which the system should answer using domain knowledge.
Due to the complex interactions between the user and the system, errors are prone to happen which in turn leads to user dissatisfaction and low ratings.
Being able to predict the rating of an interaction is thus a critical step to understand the problems of the system, and act accordingly in both an online and offline setting~\cite{offline_online_satisfaction, socialbot_big_rating_prediction}.

The aforementioned problems motivate our work in offline rating prediction, which is a challenging scenario, where the goal is to predict a rating at the end of the interaction taking into account the whole conversation.
This task helps discover patterns in user ratings and, more importantly, detect problematic conversations which can be further analyzed to discover avenues for system improvement. 
In particular, and to the best of our knowledge, we are the first to tackle the problem of rating prediction in a Conversational Task Assistant (CTA)~\cite{taskbot_overview} setting.
In Figure~\ref{fig_bad_conversation}, we show an example of a low-rated CTA conversation. This happens due to various aspects, such as ASR errors recognizing ``die hair'' instead of ``dye hair'', or fallback responses, for example, when the system is not able to answer a question.
How to use these various signals to predict the rating is one of our goals.  
To evaluate the rating prediction task, we leverage data collected during the Alexa Prize TaskBot Challenge~\cite{taskbot_overview, twiz}, comprised of real human-agent CTA interactions. 
In this setting, the users interact with Alexa devices, mainly using their voice in a conversational, multi-turn, and multimodal way.

Evaluating conversational assistants is an active and challenging research subject in which the gold-standard metric is human-based evaluation~\cite{facebook_compare_evaluation, kiseleva_understanding_user_satisfaction_2016, tod_satisfaction}.
Many works~\cite{kiseleva_predicting_user_satisfaction_2016,italian_rating_prediction, offline_online_satisfaction, socialbot_big_rating_prediction} leverage this human-labeled data to train automatic methods for conversational assistants evaluation.
For example, we highlight the design of manual features in~\cite{last_turns_doomed} for a flight booking system, and \cite{kiseleva_predicting_user_satisfaction_2016} for search dialogs.
In~\cite{italian_rating_prediction, offline_online_satisfaction, socialbot_big_rating_prediction}, models are proposed to automatically predict the rating/satisfaction on Alexa's SocialBot Challenge~\cite{socialbot_overview}.
In particular, \citet{offline_online_satisfaction} show advantages in leveraging both textual and behavioral features. Motivated by this work, we created user behavior features that are specific to the CTA setting. Moreover, we use the recent advances with the use of Transformer-models~\cite{vaswani_attention, bert_original, t5_model} to create conversational-flow features.
Despite the significant differences between chit-chat (SocialBot) and the CTA (TaskBot) setting, we believe that a combination of both types of features can bring improvements in rating prediction. With this, we combine the features into a single model which we call \textit{TB-Rater} (\textit{Transformer-Behavior Rater}) that surpasses the considered baselines.
To conclude, we perform an ablation study, showing how the various design decisions influence the model's results, and analyze the importance of the behavior features in this novel setting.

\section{Transformer-Behavior Rater}
\label{sec_approach}
In this section, we present our proposed model \textit{Transformer-Behavior Rater} (\textit{TB-rater}), which combines two sets of features.

\subsection{Model Architecture}
\label{sub_model_architecture}

\subsubsection{Conversational-Flow Features}
\label{sub_sub_text_features}
The content and flow of the dialog conveys information about the current state and rating. 
Thus, we propose to use conversational-flow features with the aim of capturing intricate and discriminative dialog flows.
To model these features computationally, we use a Transformer-based~\cite{vaswani_attention} language model, which is able to capture various patterns in the language and derive a representation of the conversation's state~\cite{simpletod, ubar}.

\noindent
We represent each turn ($T_i$) of the dialog as follows:
\begin{equation}
    T_i = ``[S] \ [RG_i] \ S_{i} \ [U] \ [INT_i] \ U_{i} " \ ,
\label{eq_turn}
\end{equation}
where $S_i$ and $U_i$ are the system and user utterances, separated by special tokens \textit{[S]} and \textit{[U]} denoting the beginning of a speaker's turn.
We go beyond the utterances and include flow-based information in the form of the intent detected $[INT_i]$, which has proven useful in~\cite{user_satisfaction_smart_speakers, iart_importance}, and the response generator selected/activated $[RG_i]$ to provide extra information to the model.

\noindent
A conversation with $n$ turns is modeled as the sequence:
\begin{equation}
    ``[CLS] \ [DEV] \ [DOM] \ T_1 \ldots T_i \ldots T_n " \,
\label{eq_text_input}
\end{equation}
The first token of the sequence is a special $[CLS]$ token~\cite{bert_original}. Specific to our model, we use additional special tokens denoting the type of device $[DEV]$ (screen/screen-less) and the domain of the user's task $[DOM]$, which can be none, a recipe, or a DIY~\cite{taskbot_overview}.
We use all turns of the conversation and perform left truncation of the input when it is over the maximum sequence length.
Finally, we use the embedding of the \textit{[CLS]} token ($emb_{[CLS]}$) as the representation of the conversation.
This first set of features is then complemented with user behavior features.

\subsubsection{Behavior Features}
\label{sub_sub_behavior_features}
Taking inspiration from~\citet{offline_online_satisfaction}, which showed a performance increase when combining text and behavior features. We follow a similar pattern and add manually engineered features specific and unique to the CTA domain, with the aim of providing more domain context to the model.
In particular, we use the last turn of the conversation ($T_n$) to get the behavior features ($B_n$).
In total, we created 70 features divided in General, System-Induced, and CTA-Specific features.

\noindent
\textbf{General.} 
Table~\ref{tab_general_features} presents general conversational features, where we can see a large overlap with the features in~\citet{offline_online_satisfaction}.

\begin{table}[tbp]
\centering
\footnotesize
\caption{General Features. A $/$ in a feature denotes $>1$ feature.}
\label{tab_general_features}
\begin{tabular}{@{}ll@{}}
\toprule
\textbf{General Feature}            & \textbf{Description}                            \\ \midrule
SessionDuration            & Duration in seconds                             \\
TurnDuration                  & Turn $i$ duration in seconds             \\
Avg/Max TurnDuration             & Avg/Max turn duration             \\
Turns                & \# Turns                                         \\
Utterance Pos/Neg\footnotemark{}               & \# User positive/negative utterances                      \\
AvgUtterance Pos/Neg                     & Avg user positive/negative utterances                    \\
Offensive/Sensitive\footnotemark{}             & \# Turns w/ offensive/sensitive content                  \\
User/System WordOverlap        &  Word overlap ratio btw consecutive user/system             \\
UserSystemWordOverlap   & Word overlap ratio btw $S_{i-1}$ and $U_i$             \\
Avg User/System WordOverlap  & Avg Word overlap ratio btw user/system             \\
AvgUserSystemWordOverlap & Avg Word overlap ratio btw user and system             \\
Words User/System           & Total \# words in $U_i$ / $S_i$                           \\
AvgWords User/System       & Avg \# words in user/system                            \\
Unique User/System Words          & Unique words btw consecutive user/system turns   \\
\bottomrule
\end{tabular}%
\end{table}

\begin{table}[tbp]
\centering
\footnotesize
\caption{CTA-specific. A $/$ in a feature denotes $>1$ feature.}
\label{tab_taskbot_features}
\begin{tabular}{@{}p{0.50\linewidth}l@{}}
\toprule
\textbf{CTA-Specific Feature} & \textbf{Description}               \\ \midrule
StepsRead               & \# Steps read                      \\
Repeated User/System Utterance       & \# Repeated user/system utterances        \\
Resumed                  & User resuming session              \\
HasScreen                     & User is using a device w/ screen   \\
Screens             & \# Screens visited                               \\
Searches             & \# Search requests                               \\
RepeatedSearches & \# Repeated search requests                      \\
ResultPages         & \# Result pages seen                             \\
Started/Finished Task                & User started/finished a task                    \\
FallbackExceptions         & \# System fallback responses                     \\
Domain                        & Domain of task (recipe, DIY, none) \\
Curiosities Accepted/Denied             & \# Curiosities user accepted/rejected       \\
CuriositiesSaid                 & \# Curiosities the system said        \\ \midrule
\textbf{Phase-based}                      &                                    \\ \midrule
Greeting/Search/Task Overview/ Ingredients/Steps/Step's Detail/Conclusion                   & \# Turns in a particular phase         \\
\midrule
\textbf{Intent-based}                      &                                    \\ \midrule
Search/None of These/Cancel/Yes/No/ Ingredients/Start Cooking/Start Steps/Next/ Next Step/More Detail/Terminate Task/ Help/Repeat/Fallback                      & \# Intents of particular type                  \\
\bottomrule
\end{tabular}%
\end{table}

\addtocounter{footnote}{-1}\footnotetext{We created a threshold-based method to identify positive/negative utterances based on an internal Amazon Alexa algorithm that uses the audio of the utterance.}
\addtocounter{footnote}{+1}\footnotetext{Offensive and sensitive content is identified by an internal Amazon Alexa classifier and by matching with a list of special words.}

\noindent
\textbf{System-Induced.} We consider the values for a particular turn, the avg, and the max across the conversation for user latency, system latency, and scores given by the ASR model, as in~\cite{offline_online_satisfaction}.

\noindent
\textbf{CTA-Specific.} 
In Table~\ref{tab_taskbot_features}, we propose CTA-specific features, such as the number of searches or steps read, the number of turns in a phase, which indicates the depth the user is going into the conversation, or the counts of a specific intent as predicted by another model.
These features were designed based on real-world interactions and thus can serve as a basis for other works in this setting.

\subsubsection{Features Combination}
\label{sub_sub_combined_features}
First, we use two feed-forward neural networks (FFNN), $FFNN_T$ and $FFNN_B$ (with ReLu activations), that take as input the $emb_{[CLS]}$ and the behavior features $B_n$, respectively.
After this, the resulting representations are concatenated and passed through a final $FFNN_{TB}$ that combines the two streams:
\begin{equation}
    FFNN_{TB}( FFNN_T(emb_{[CLS]}) \oplus  FFNN_B(B_n) ).
    \label{eq_rating_feedforward}
\end{equation}
With this approach, we make predictions benefiting from both information streams, as shown in~\cite{offline_online_satisfaction, dst_wide_features} in different domains. 
The model is then trained using the cross-entropy loss.

\section{Experiments}
\label{sec_experiments}

\subsection{Experimental Setting}
\label{sub_experimental_setting}

\subsubsection{Alexa Prize TaskBot Dataset}
\label{sub_sub_taskbot_dataset}
To evaluate our models, we use internal data collected in the first Alexa Prize TaskBot challenge~\cite{taskbot_overview,twiz}.
This challenge focuses on developing a CTA that helps users perform real-world manual tasks in the cooking and DIY domains. It is also the first multimodal challenge of this type, combining both voice-only and voice-and-screen interactions.
In this setting, the system interacted with thousands of users, and for each conversation, at the end of the interaction, they are asked to provide an optional rating on a 1 to 5 scale.
However, only about 10\% of the users provide a rating, making it hard to pinpoint which conversations require more attention, further motivating our work.
 
We used a stable version of the system to collect ratings and considered only rated conversations with a minimum of 3 turns. 
In total, we used 1681 conversations which we separated into training (90\%), validation (10\%), and test (10\%) sets. 
The statistics of the dataset are in Table~\ref{tab_taskbot_dataset_stats}.
We observe that, on average, a dialog has 8 to 9 turns, with a standard deviation of 6.8, 
indicating a large variety of conversation lengths.
In terms of the ratings, we see a larger concentration in 1 and 5, with a standard deviation of 1.55, indicating that the users generally have a strong opinion about the system's performance, as also noticed in the SocialBot domain~\cite{italian_rating_prediction}.

\begin{table}[tbp]
\centering
\caption{Alexa TaskBot Dataset statistics.}
\label{tab_taskbot_dataset_stats}
\small
\begin{tabular}{@{}lccc@{}}
\toprule
                 & \textbf{Train} & \textbf{Validation} & \textbf{Test} \\ \midrule
\# Conversations & 1344           & 168                 & 169           \\
\# Turns         & 12784          & 1390                & 1567          \\
Avg \# Turns     & 9.5            & 8.3                 & 9.2           \\ \midrule
Rating 1         & 263 (19.6\%)   & 34 (20.2\%)         & 30 (17.8\%)   \\
Rating 2         & 158 (11.8\%)   & 19 (11.3\%)         & 23 (13.6\%)   \\
Rating 3         & 179 (13.3\%)   & 23 (13.7\%)         & 24 (14.2\%)   \\
Rating 4         & 231 (17.2\%)   & 26 (15.5\%)         & 30 (17.8\%)   \\
Rating 5         & 513 (38.2\%)   & 66 (39.3\%)         & 62 (36.7\%)   \\ \bottomrule
\end{tabular}%
\end{table}

\subsubsection{Task and Metrics}
\label{sub_sub_task_and_metrics}
In this work, we define the task of rating prediction at the end of the interaction. 
This makes this task challenging due to the need for a model capable of understanding the entire conversation, and identify the non-trivial subtleties that contribute to the rating.

Following a similar approach to~\citet{offline_online_satisfaction},
we use a binary classification task by separating ratings 1-3 into 0 and 4-5 into 1, instead of using the original 1-5 rating scale.
In terms of metrics, we considered accuracy (Acc), precision (P), recall (R), and F1.

\subsection{Methods and Baselines}
\label{sub_methods_baslines}

\noindent
\textbf{Behavioral-only} - we tested the following methods \textit{RandomForest}~\cite{random_forest}, \textit{AdaBoost}~\cite{adaboost}, \textit{Bagging}~\cite{bagging},  \textit{GradientBoosting}~\cite{gradient_boosting}, \textit{XGBoost}~\cite{xgboost}, and \textit{LogisticRegression}. All methods are implemented using \textit{sklearn}~\cite{scikit_learn} and use the behavior features of the last turn.

\noindent
\textbf{Conversational-Flow-only} - To encode the dialog features, we used a BERT model~\cite{bert_original} with a classification head. 
We also adapted a T5~\cite{t5_model} model for classification. 

\noindent
\textbf{Conversational-Flow and Behavior} - we implemented \textit{ConvSat}~\cite{offline_online_satisfaction}, which combines text features at an utterance and character levels using BiLSTMs~\cite{lstm_original}, which are combined with behavioral features. 
We also present the results of the proposed \textit{TB-Rater} model.\footnote{Code available at \url{https://github.com/rafaelhferreira/cta_rating_prediction}}

\subsection{Results}
\label{sub_results}

\subsubsection{General Results}
\label{sub_sub_general_results}
We present the results of the various methods on the Alexa TaskBot Dataset in Table~\ref{tab_rating_prediction_results}. 
First, we observe that the best behavior-only method is the \textit{SVM}.
Regarding conversational-flow-only methods, the \textit{BERT-Base} model achieves the best results, surpassing the enc-dec model \textit{T5}. This might be explained by BERT having a specific and pre-trained classification token~\cite{bert_original}, while T5 is adapted to classification using a text-to-text paradigm~\cite{t5_model}.
The \textit{BERT-Base} approach also surpasses the best behavior-only method (\textit{SVM}), showing that only using conversational-flow information may be a good alternative for rating prediction,
avoiding the need for the design of domain-specific features.
Comparing the conversational-flow and behavior models, we see that the best results are achieved by the proposed \textit{TB-Rater} model, surpassing all of the considered baselines. 
This result is in line with previous work~\cite{offline_online_satisfaction, dst_wide_features} that showed advantages in combining text and behavior features.
However, \textit{ConvSat}~\cite{offline_online_satisfaction}, which also uses both types of features, did not perform as well. We believe this may be due to having a small amount of training data to effectively train the character and word level embeddings, making Transformer-based models a more robust approach.
To conclude, the results show that it is possible to have conversation-flow-only models that are on par with classic approaches based on manually engineered features. We also show that combining both types of features in \textit{TB-Rater} brings an improvement in performance.

\begin{table}[t]
\centering
\small
\caption{Avg. result of 3 runs on the Alexa TaskBot test set.}
\label{tab_rating_prediction_results}
\begin{tabular}{@{}lcccc@{}}
\toprule
\textbf{Method}    & \textbf{Acc}    & \textbf{P}     & \textbf{R}     & \textbf{F1}    \\ \midrule
\multicolumn{5}{l}{\textbf{Behavior-Only}}                                             \\ \midrule
RandomForest       & 66.3 & 66.0 & 65.8 & 65.8 \\
AdaBoost           & 64.5 & 64.2 & 63.7 & 63.7  \\
Bagging            & 67.1  & 66.8 & 66.8 & 66.8 \\
GradientBoosting   & 66.3  & 66.0 & 66.0 & 66.0 \\
XGBoost            & 64.5 & 64.2 & 64.1 & 64.1  \\
LogisticRegression & 67.5 & 67.5 & 66.4 & 66.4  \\
SVM                & {\ul 68.6} & {\ul 68.4} & {\ul 68.2} & {\ul 68.3} \\ 
\midrule
\multicolumn{5}{l}{\textbf{Conversational-Flow-Only}}                                                 \\ \midrule
BERT-Base & {\ul 69.0}    & {\ul 69.2}    & {\ul 69.1}    & {\ul 68.8}    \\
T5-Base            & 66.9 & 67.2 & 67.1 & 66.8 \\ \midrule
\multicolumn{5}{l}{\textbf{Conversational-Flow and Behavior}}                                         \\ \midrule
ConvSat~\cite{offline_online_satisfaction} &     63.4 &    61.9  &    61.4  &      63.3 \\
TB-Rater (Ours)  & \textbf{69.6} & \textbf{70.0} & \textbf{70.0} & \textbf{69.6} \\ \bottomrule
\end{tabular}%
\end{table}

\subsubsection{Ablation Study}
\label{sub_sub_ablation_studies}

In Table~\ref{tab_ablation_results}, we analyze how our design decisions influence the model's results.
As we saw previously, removing behavioral features negatively affects the results. 
In \textit{w/o Step Token}, we keep the text of the task's step instead of replacing it with a special token $[STEP]$. We see a decrease in performance, which we attribute to the step text not being especially important for the rating. Adding to this, keeping the text of a step also decreases the number of turns inputted into the Transformer model due to steps typically being long.
In \textit{w/o Additional Tokens}, we remove the special tokens pertaining to the device, domain, intent, and response generator, but we keep the special $[STEP]$.
Again, we see that adding extra information in the form of these tokens increases performance.
Finally, we test the \textit{TB-Rater} model but truncate inputs larger than the maximum input size from the \textit{right side} (end of the conversation) instead of the left side.
Here, we observe the worse results out of all methods. 
This result shows that focusing on the end of the conversation 
is more important to predict the rating, this can be attributed to the last turns having more impact than the ones at the beginning, indicating a possible recency bias.

\begin{table}[tbp]
\centering
\small
\caption{\textit{TB-Rater} ablation study on Alexa TaskBot test set.}
\label{tab_ablation_results}
\begin{tabular}{@{}lcccc@{}}
\toprule
\textbf{Method}     & \textbf{Acc} & \textbf{P} & \textbf{R} & \textbf{F1} \\ \midrule
TB-Rater  & \textbf{69.6} & \textbf{70.0} & \textbf{70.0} & \textbf{69.6}            \\ \midrule
w/o Behavior (i.e., BERT BASE)   &  69.0    & 69.2    & 69.1    & 68.8    
 \\
w/o Step Token &     66.7       &   68.0         &    67.3        &      66.3       \\
w/o Additional Tokens     &    65.9          &    66.4         &        66.0     &      65.5       \\
Right Side Truncation       &    63.9          &    64.2         &      64.2        &   63.9  \\ \bottomrule
\end{tabular}%
\end{table}

\subsubsection{Error Analysis}
\label{error_analysis}
While user subjectivity plays an important role~\cite{italian_rating_prediction, socialbot_big_rating_prediction}, we believe that a portion of the model's errors can be categorized. Thus, we analyze \textit{TB-Rater}'s 50 error cases (counts of error types are given between parentheses).
We noticed that the model generally gives a low rating if the interaction is stopped early, but the user is able to find and/or start a task (12). Another mistake is when the user starts a task that is different from the one the user is looking for but still goes further into the task, usually with consecutive dull responses (e.g., next step). In this case, the model predicts a high rating despite the user giving a low one (9).
There were also cases where despite the system giving a considerable number of fallback answers, the conversation still moves forward, however, the model predicts this as an unsatisfactory conversation (10).
Finally, user ratings have a lot of variability, and some do not seem to reflect how the interaction went, for example, ``throw-away''/bad interactions that returned high ratings (7), or interactions where the user is not impressed with the system, returning a low rating despite the system responding to every request correctly (12).
These results reaffirm the volatility of user ratings~\cite{italian_rating_prediction, offline_online_satisfaction} and the difficulty of the task, shedding light on the most common error cases.

\subsubsection{Behavior Feature Importance}
\label{sub_sub_behavior_importance}
\begin{figure}[tb]
    \centering
    \includegraphics[trim={0 5mm 0 3mm},clip,width=0.84\linewidth]{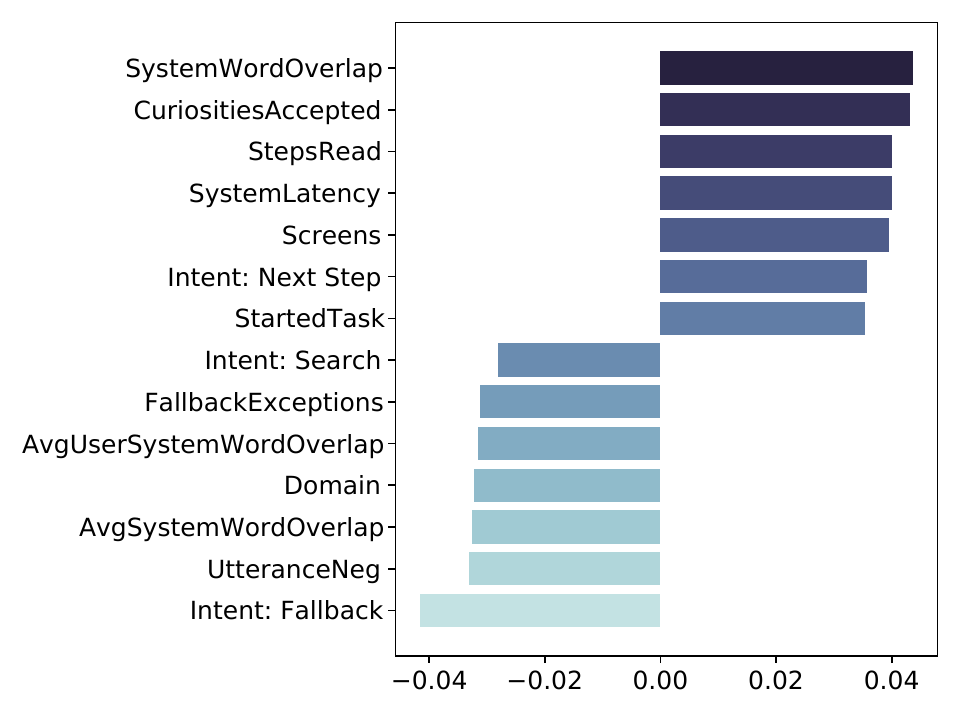}
    \caption{Logistic Regression Top-14 absolute 
    coefficients.}
    \label{fig_behavior_feature_importance}
\end{figure}

In Figure~\ref{fig_behavior_feature_importance}, we present the top-14 abs. feature coefficients for the \textit{Logistic Regression} model.  
Here positive/negative scores indicate a feature that predicts a positive/negative rating.
Starting with the \textit{system word overlap} on the last turn, this indicates that the last two system utterances share a large number of words.
This feature is relevant because when the user finishes a task there is a large token overlap. 
The higher \textit{system latency} on the last turn also appears to have importance in a positive rating, which at first seems counter-intuitive. After a closer analysis, we attribute this to the last turn of a finished task having a larger latency while an abrupt stop has a latency value of zero.
In practice, these two features indicate that finishing a task is an important signal for predicting the rating.
Other features such as the number of \textit{steps read}, \textit{next step}, and \textit{started task} suggest that the user is engaged with the system and going deeper into a task.

Regarding the negative coefficients, we see that a larger number of \textit{fallbacks}
leads to a lower rating. 
The \textit{average system overlap} denotes that the system is saying a similar response in multiple turns, which might indicate that the user is stuck.
Finally, a higher value of \textit{domain} indicates that the user did  
not search for a task, and in opposition, a high \textit{number of searches} indicates that the user is struggling to find a task, resulting in a lower rating.
It is also worth noting that out of the 14 features, 9 are from the CTA-specific set, showing the relevance of the proposed features.

\section{Conclusion}
\label{sec_conclusion}
In this paper, we propose \textit{TB-Rater}, a model that combines conversational flow and behavioral features to perform rating prediction in the novel CTA setting.
We show the advantages of combining both types of features by evaluating on human-agent interactions collected in the Alexa TaskBot challenge.
Moreover, we provided a comprehensive set of CTA-specific features and measured their importance.
The model proposed can be used to estimate a rating, which may allow for the discovery and prioritization of system errors. 
In future work, we intend to apply the model in an online setting, using its predictions to change the course of a conversation. 

\begin{acks}
This work has been partially funded by the Amazon Science - TaskBot Prize Challenge 2021, by the NOVA LINCS project Ref. UIDP/04516/2020, by FCT Ref UI/BD/151261/2021, and by the CMU| Portugal iFetch project LISBOA-01-0247-FEDER-045920.
\end{acks}




\bibliographystyle{ACM-Reference-Format}
\balance
\bibliography{sample-base}

\appendix


\end{document}